%% file: iclr2025_conference.tex
\documentclass{article} 
\usepackage{iclr2025_conference,times}

\input{math_commands.tex}

\usepackage{hyperref}
\usepackage{url}
\usepackage{subfigure}
\usepackage{comment}
\usepackage{xcolor}
\usepackage{subcaption}

\usepackage{caption}    %
\usepackage{booktabs} %
\usepackage{float}
\usepackage{graphicx}
\usepackage{tcolorbox}
\usepackage{listings}
\usepackage{xcolor}
\definecolor{lightgray}{gray}{0.9}
\title{FastRM: An efficient and automatic explainability framework for multimodal generative models}

\author{Gabriela Ben Melech Stan, Estelle Aflalo\thanks{ Equal Contribution.}, Man Luo, Shachar Rosenman, Tiep Le, Shao-Yen \\ \textbf{Tseng, Vasudev Lal}  \\
Intel Labs, Santa Clara \\
\texttt{\{gabriela.ben.melech.stan, estelle.aflalo, man.luo, }\\ \texttt{shachar.rosenman, tiep.le, shao-yen.tseng, vasudev.lal\}@intel.com} \\
\And
Sayak Paul \\
Hugging Face, India
\texttt{\{sayak\}@huggingface.co} \\
}

\iclrfinalcopy 
\begin{document}

\maketitle
\begin{abstract}
Large Vision Language Models (LVLMs) have demonstrated remarkable reasoning capabilities over textual and visual inputs. 
However, these models remain prone to generating misinformation.
Identifying and mitigating ungrounded responses is crucial for developing trustworthy AI. Traditional explainability methods such as gradient-based relevancy maps, offer insight into the decision process of models, but are often computationally expensive and unsuitable for real-time output validation. 
In this work, we introduce FastRM, an efficient method for predicting explainable Relevancy Maps of LVLMs. Furthermore, FastRM provides both quantitative and qualitative assessment of model confidence. Experimental results demonstrate that FastRM achieves a 99.8\% reduction in computation time and a 44.4\% reduction in memory footprint compared to traditional relevancy map generation.
FastRM allows explainable AI to be more practical and scalable, thereby promoting its deployment in real-world applications and enabling users to more effectively evaluate the reliability of model outputs.
\end{abstract}

\section{Introduction}

Large Vision Language Models (LVLMs) have emerged as the next family of powerful foundation models that further advance AI applications. 
 LVLMs have shown promising performance enhancements in a wide range of applications such as healthcare \cite{huang2023chatgpt, mesko2023impact, li2024llava}, autonomous-driving systems \cite{liao2024gpt, cui2024survey}, education \cite{bewersdorff2024taking}, and virtual assistants \cite{team2023gemini, 4o}.
However, despite their impressive capabilities, LVLMs are often constrained due to the opaqueness of their decision-making process and susceptibility to generating hallucinations, describing cases when the model produces responses that are not supported by any information given in the input, but instead are drawn from the language priors within the LLM \cite{li-etal-2023-evaluating, liu2024mitigating, bai2024hallucination}. 
To address this challenge, explainability methods have sought to provide insight into the reasoning behind responses as a means to validate model outputs and mitigate future hallucinations.

Some common approaches to explaining model decisions include gradient-based visualization methods like Grad-CAM \cite{selvaraju2017grad} and relevancy maps \cite{chefergame}. 
These methods have proven to be very useful in providing insight into the generation process of LVLMs \cite{lvlm, zhang2024redundancy}, as well as improving visual aptitude \cite{chefer2022optimizing}.  

Gradient-based relevancy maps are valuable for understanding model predictions but are computationally expensive and memory-intensive, especially for large models with long output sequences, making deployment in real-time or resource-constrained scenarios challenging.
To tackle these challenges, we introduce a novel probing module for LVLMs, enabling the prediction of explainable relevancy maps without reliance on gradient-based methods.
This approach seeks to distill the explainability properties of relevancy maps into a proxy model. Such inherently explainable model outputs are crucial for ensuring the reliability and safety required for the broad adoption of LVLMs. 

Our contributions are threefold:
(1) We introduce a novel approach to enhancing model interpretability by distilling insights from a high-performance explainability method into a simpler alternative with lower computational and memory demands. 
This approach differs from existing techniques that distill the complex model into an inherently interpretable one.
(2)  Our model is able to provide on-the-fly explanations that can be used to judge the validity of model decisions dynamically, enhancing its practicality and making it more deployable in real-world scenarios.
(3) We introduce a novel metric to quantify model uncertainty based on the entropy of outputs of our model.


\section{Related Work}
\paragraph{Explainable methods}
In their study, ~\cite{xaimethods} introduce three dimensions to classify explainable methods: model distillation, intrinsic methods, and visualization methods. Model distillation such as ~\cite{lime} offers a way to provide explanations alongside the model's outputs by mimicking the input/output behavior. In general, the distillation method is interpretable by nature.
Intrinsinc methods (such as  ~\cite{intrinsic}) make use of a model that has been specifically created to render an explanation along with its output. An additional explanation “task” can be added to the original model and jointly trained along with the original task providing the expected explanation. Here, the explainable method is by itself a black box but helps to understand the model behaviors. 
 Lastly, visualization methods  identify input features causing a maximum response influencing the model’s output. Methods include perturbation-based and gradient-based techniques (~\cite{Mahendran_2016,dabkowski2017realtimeimagesaliency,simonyan2014deepinsideconvolutionalnetworks}). Our work positions itself at the intersection of the 3 types of methods enunciated above. We propose to extend a black box LVLM to render a visual explanation by distilling from interpretable methods.

\paragraph{Relevancy Map Based Interpretability}
Several methods have been proposed to actively guide models by regularizing their attributions across various tasks, including classification ~\cite{ross2017rightrightreasonstraining, gao2022aligningeyeshumansdeep, Gao_2022}, visual question answering ~\cite{selvaraju2019takinghintleveragingexplanations, teney2020learningmakesdifferencecounterfactual}, segmentation ~\cite{li2018telllookguidedattention}, and knowledge
distillation ~\cite{fernandes2022learningscaffoldoptimizingmodel}.
These methods aim not only to enhance performance but also to guarantee that the model is `right for the right reasons' ~\cite{ross2017rightrightreasonstraining}.
For transformer-based architectures, relevancy maps have proven to be effective tools for providing interpretable explanations of model predictions ~\cite{chefer2021genericattentionmodelexplainabilityinterpreting, zhang2024forward, stan2024lvlminterpretinterpretabilitytoollarge}.
These explainability techniques play a crucial role in elucidating model decision-making processes, often revealing that models may disregard certain objects ~\cite{chefer2021transformerinterpretabilityattentionvisualization} or fail to attend to relevant input regions \citet{lvlm}. The approach proposed by \citet{chefergame} utilizes attention maps from each attention layer along with their gradients to generate relevancy maps. However, this method incurs significant memory overhead and increases inference latency, rendering it less suitable for resource-constrained environments.

\paragraph{Confidence Score}
Previous methods for estimating model confidence include entropy-based measures applied to the models output probabilities, often requiring calibration due to sensitivity to distribution shifts \cite{tu2024empiricalstudymatterscalibrating, yona2022usefulconfidencemeasuresmax}. Alternatively, consistency-based assessment methods, which evaluate model responses to semantically equivalent question rephrasing  \cite{khan2024consistencyuncertaintyidentifyingunreliable}, or vision perturbation \cite{zhang2024vluncertaintydetectinghallucinationlarge}, require multiple inferences through the LVLM. \cite{fang2024uncertaintytrustenhancingreliability} introduced a method to assess visual uncertainty of each image token by projecting it into the text space. Our method offers an efficient way to assess model confidence both qualitatively through the generated relevancy maps, increasing interpretability and quantitatively through the computation of their entropy.

\section{Method} 
\begin{figure}[thb]
    \centering
    \begin{minipage}{0.9\textwidth}
        \centering
        \includegraphics[width=\columnwidth]{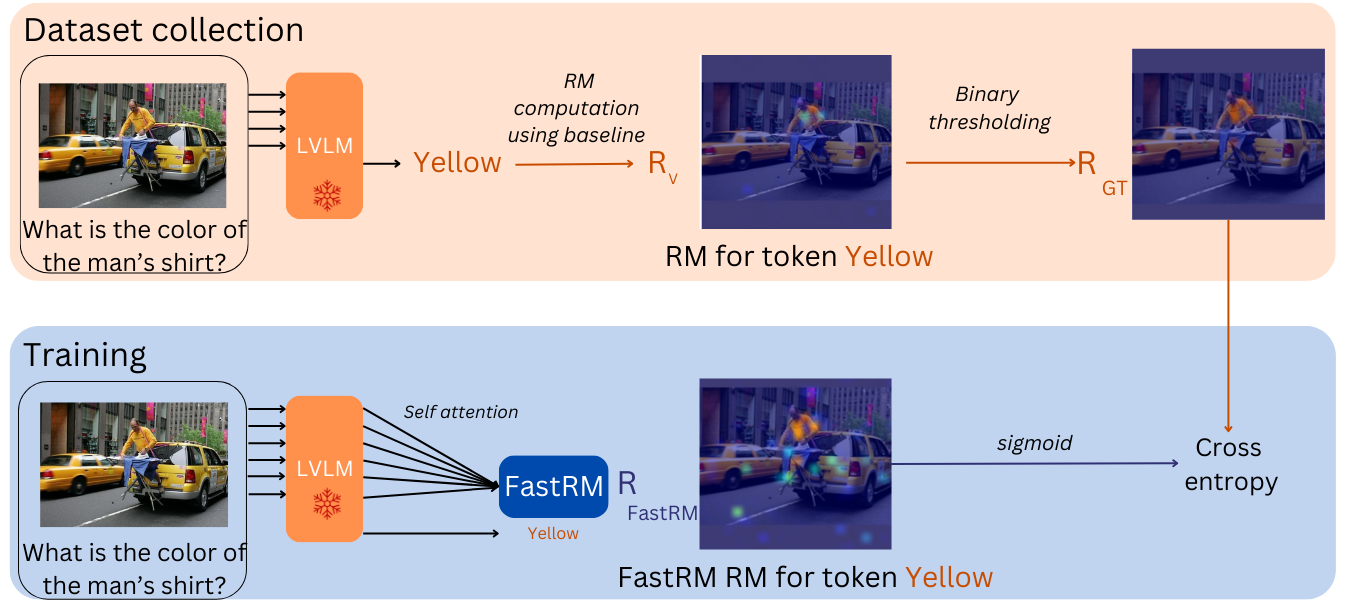}
        \caption{Overview of FastRM. Given an input, the LVLM produces hidden states for each generated token. Then, FastRM generates the relevancy map $R_{FastRM}$ which is subsequently compared to $R_{GT}$ obtained after binarizing the baseline relevancy map. }
        \label{fig:model_diag}
    \end{minipage}\hfill
\end{figure}

In this work, we introduce FastRM, a framework for efficiently predicting relevancy maps that highlight image regions attended to by a model during response generation.
Specifically, FastRM features a lightweight proxy module that mimics the overall behavior of the relevancy maps introduced by \cite{chefergame}, which we denote as the Baseline throughout this paper. 
An high-level overview of FastRM is shown in Figure \ref{fig:model_diag}.
Unlike conventional methods that require storing all attention weights and computing their gradients with respect to model outputs, our model offers a fast way to estimate the relevancy map with a reduced memory footprint and latency.

\paragraph{Dataset Creation}
To train our FastRM models, we utilized the VQA training dataset \cite{vqa}. For each input question, we generated the answer and for each output token, we save both the relevancy map $R$, as computed in \cite{chefergame}, and the final hidden states $H$.

Given that the input sequence consists of \( N_{in} \) tokens and the output sequence consists of \( N_{out}\) tokens. The total length is thus \(N = N_{in} + N_{out}\). We denote the relevancy scores matrix of all tokens up to output token \(i\) as \( R_i \) with dimensions \( (N_{in}+i, N_{in}+i) \). For example, \( R_0 \) will represent the relevancy scores of all input tokens and is of shape \( (N_{in}, N_{in}) \).

Each row \(j\) of the relevancy matrix represents the relevancy scores of token \(j+1\) with respect to all previous \(j\) tokens. We denote this vector of length \(j\) as the relevancy map of token \(j+1\). Specifically,  the last row of \( R_i \), \( R_i[-1,:] \), will represent the relevancy map of the output token \(i\) to all previous input and output tokens \( (N_{in}+i\).
Since we focus on the relevance of the output tokens to the visual input modality, we utilize only the relevancy scores associated with the input vision patches of each relevancy map. For LLaVA types of architecture, the number of patches is 24x24=576. This results in an output of size $(1, 576)$.

\begin{figure}[h]
\centering  
\resizebox{0.8\columnwidth}{!}{%
\subfigure[Relevancy distribution]{\label{fig:Distribution-FastRM-7}\includegraphics[height=50mm]{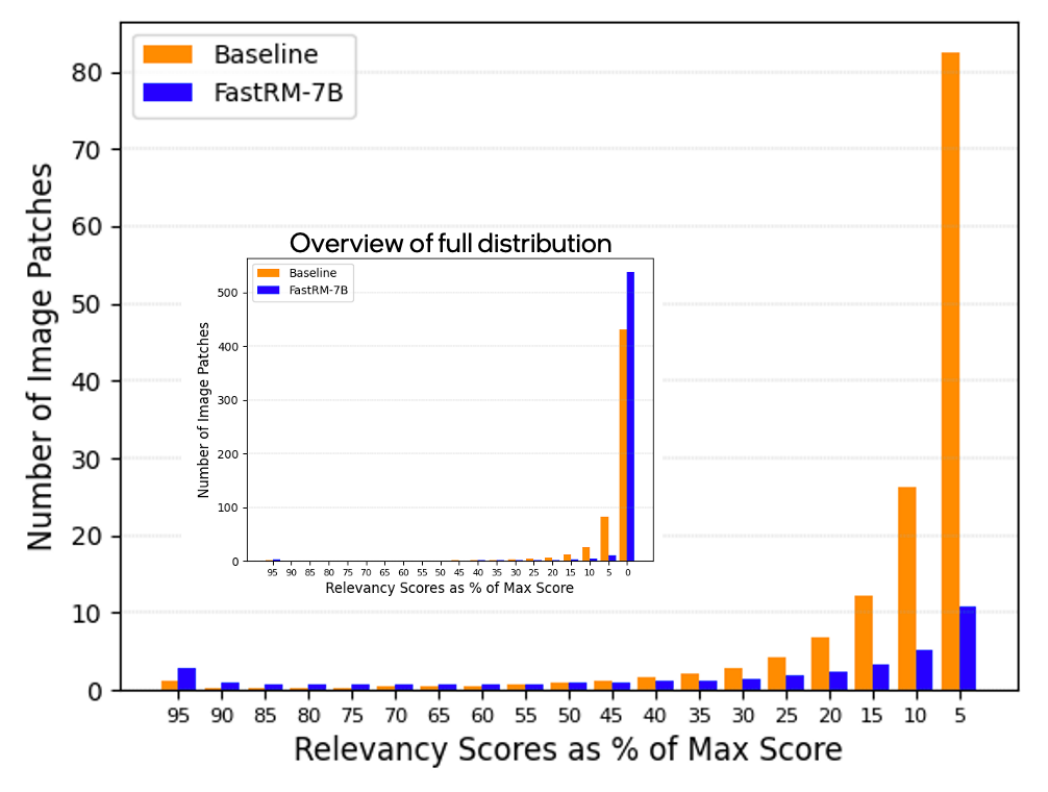}}
\subfigure[Output Lengths]{\label{fig:histo_len}\includegraphics[height=50mm]{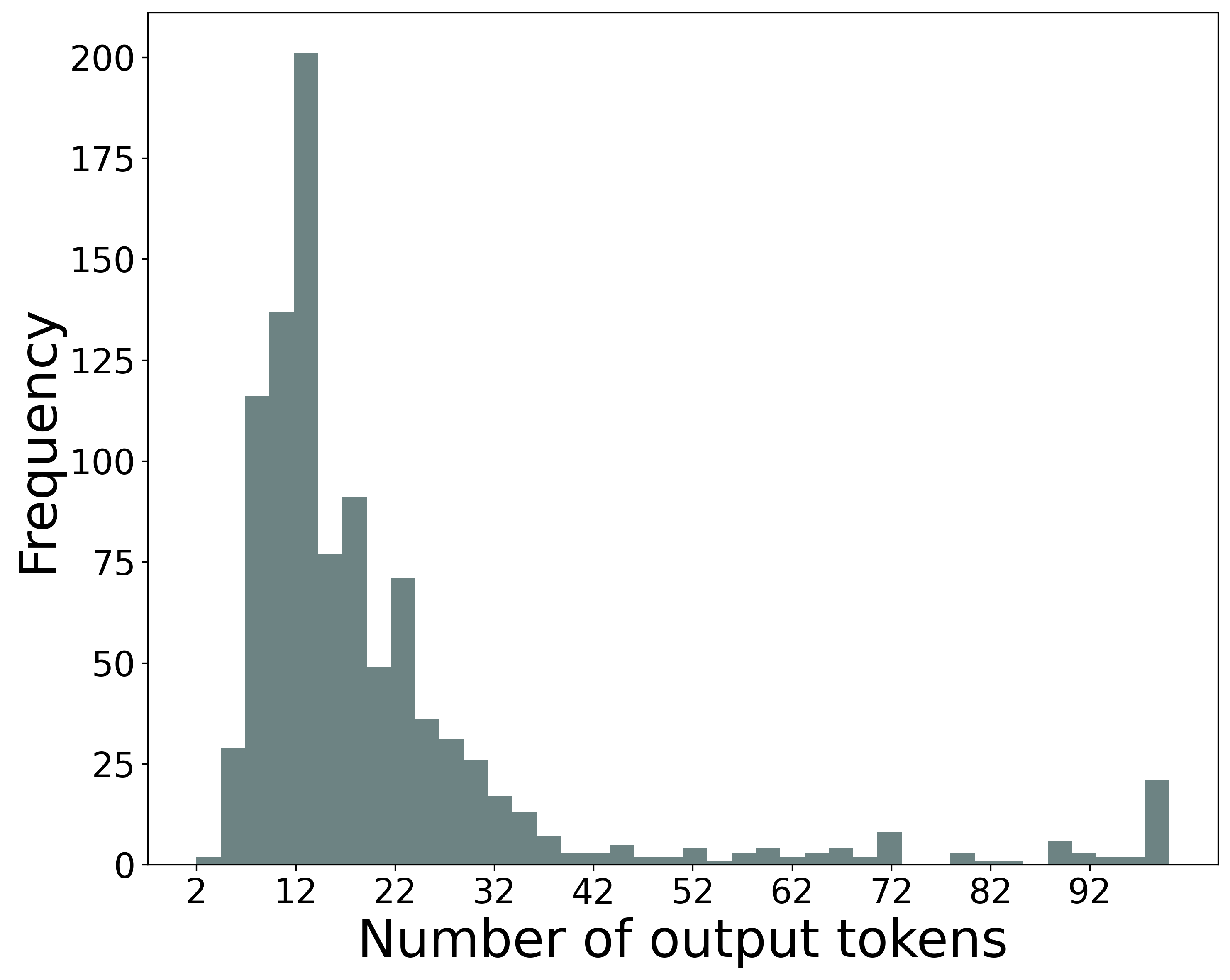}}
}
\caption{Left: Distribution of image patches relevancy scores for the baseline and FastRM-7. Showing patches with relevancy scores exceeding 5\% of the maximum relevancy value for each image, while the inset shows the full distribution. Right: shows the distribution of the output lengths.}
\end{figure}

Since over 90\% of image tokens have relevancy scores below 5\% of the maximum, as shown in Figure \ref{fig:Distribution-FastRM-7}, this means that fewer than 10\% have significant scores. This highlights the need to focus only on the most relevant tokens, as lower-scored ones are often background noise. To address this, we applied a threshold to binarize our input, referred to as ``labeling threshold", aiming to prioritize the prediction of the image's relevant portions while minimizing emphasis on less significant areas.

For a relevancy map associated with a token at position \textit{j}, the labeling threshold is applied as follows:\\
\begin{equation}
R_{GTj} =
\begin{cases} 
1 & \text{if } R_{[-1, j]V} \geq \max(R_{[-1, j]V}) \times \text{threshold} \\
0 & \text{otherwise}
\end{cases}
\label{eq:thresholdeq}
\end{equation}

Where $R_v$ is the relevancy with respect to the visual tokens, while the ``labeling threshold" is defined empirically and set to 0.3 (see ablations below). As the labeling threshold increases, fewer pixels are considered during training, causing the maps to focus on smaller but more significant regions of interest. Figure \ref{fig:model_diag} shows the data collection pipeline. Our training dataset contains 100,000 samples, representing 10,000 queries from VQA, with each query averaging 10 tokens per response.

\paragraph{FastRM Model}
 FastRM consists of a normalization layer followed by a single-head self-attention layer. FastRM processes the last hidden states of an LVLM to predict the corresponding visual relevancy map.
 It employs an attention layer to compute and output the attention scores. We formulate the task as a classification problem for each image patch. Our model predicts the probability of a patch being relevant to the model decision.
Our approach depends solely on the representation of the final hidden state. Since the original implementation of the relevancy maps from \cite{chefergame} places more emphasis on the attention scores of the last hidden states, we hypothesize that the last hidden state, derived from these attention scores, would contain sufficient information to produce these maps.
We can thus save memory and only use the last hidden states instead of storing all of the attentions and their gradients during runtime. A diagram of the model is shown in Figure \ref{fig:model_diag}.


FastRM was trained on a dataset of 100k samples, where each sample is a tuple of (last hidden state, vision relevancy vector) as described above. First, the last hidden state was normalized, then passed through FastRM, generating attention scores. We then extract the scores of the current generated token with respect to the image tokens that we processed with a sigmoid activation. 
For training, We used the Adam optimizer with a learning rate of $2 \times 10^{-5}$, and a batch size of 128. The training was conducted for 3500 steps using cross-entropy loss.


\section{Results} 

\begin{figure}[h]
    \centering
    \begin{minipage}[c]{0.38\textwidth}
        \centering
        \begin{lstlisting}[language=Python,basicstyle=\scriptsize]
model = LVLM.from_pretrained()
proxy = FastRM()

out = model.generate(inputs)
for lhs in out.last_hidden_states:
     rm = proxy(lhs)
        \end{lstlisting}
        \caption{Python code snippet}
        \label{fig:code}
    \end{minipage}%
    \hfill
    \begin{minipage}[c]{0.5\textwidth}
        \centering
        
        \includegraphics[width=\linewidth]{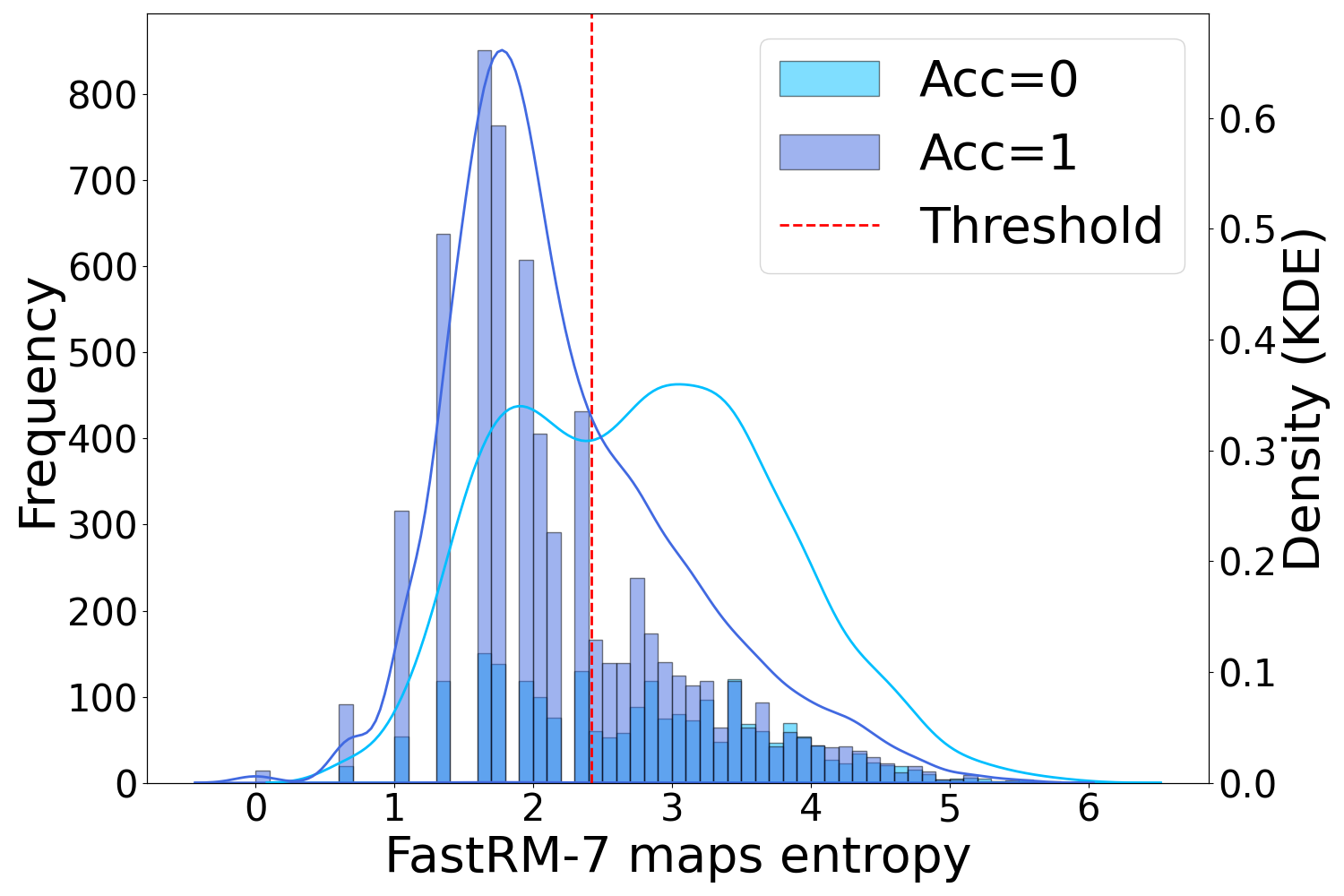}
        \caption{Entropy as a measure of uncertainty}
        \label{fig:histo_uncertainty}
    \end{minipage}
\end{figure}

Relevancy maps highlight the image regions a model focuses on when generating an answer, improving interpretability and trust. FastRM enhances this by providing an efficient and easy-to-use method for generating relevancy maps during inference and interpreting decision-making, see code snippet Figure \ref{fig:code}.
When the model attends to the relevant region of the image, we can have greater confidence in the accuracy of the answer. This hypothesis will be measured quantitatively in this section.
We showcase the validity of FastRM on the LLaVA family of LVLMs \cite{llava} and apply our method on the variants LLaVA-v1.5-7B and LLaVA-v1.5-13B to produce models which we name FastRM-7 and FastRM-13, respectively. 
The difference between the FastRM-7 and FastRM-13 architectures lies in the hidden sizes of the attention layer, which are 4096 and 5120, respectively. 
Experiments will be carried out for 10,000 randomly selected samples from VQA validation set. 
In this section, we provide both quantitative and qualitative evaluations, along with ablations studies.

\paragraph{Quantitative} 
We computed the accuracy and F1 score of FastRM by comparing the binarized labels with the predicted output obtained by applying different classification thresholds. 
The labeling and classification thresholds are distinct and serve different purposes. The labeling threshold is employed during the data preprocessing stage to define the relevance of data points, directly influencing the ground truth labels the model learns from. Changes in this threshold alter the training examples. In contrast, the classification threshold is set at inference time to determine how the model's probability score output is converted into a final prediction to decide whether the image token is relevant or irrelevant. Optimizing the classification threshold enables achieving the highest F1 score, without modifying the underlying model.


\begin{figure}[h]
\centering  
\resizebox{0.5\columnwidth}{!}{%
\subfigure[Accuracy]{\label{fig:accuracy}\includegraphics[width=50mm]{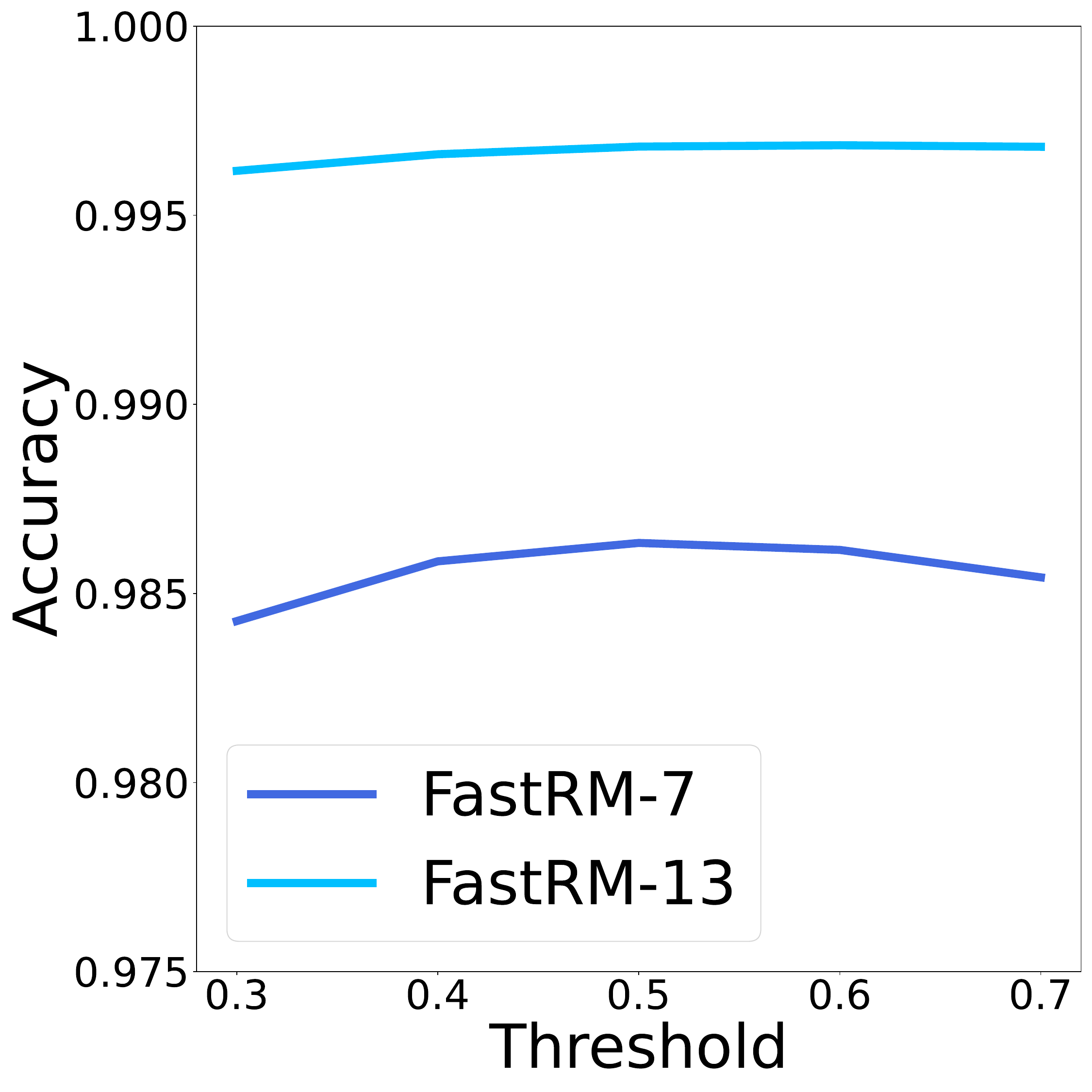}}
\subfigure[F1 score]{\label{fig:f1}\includegraphics[width=50mm]{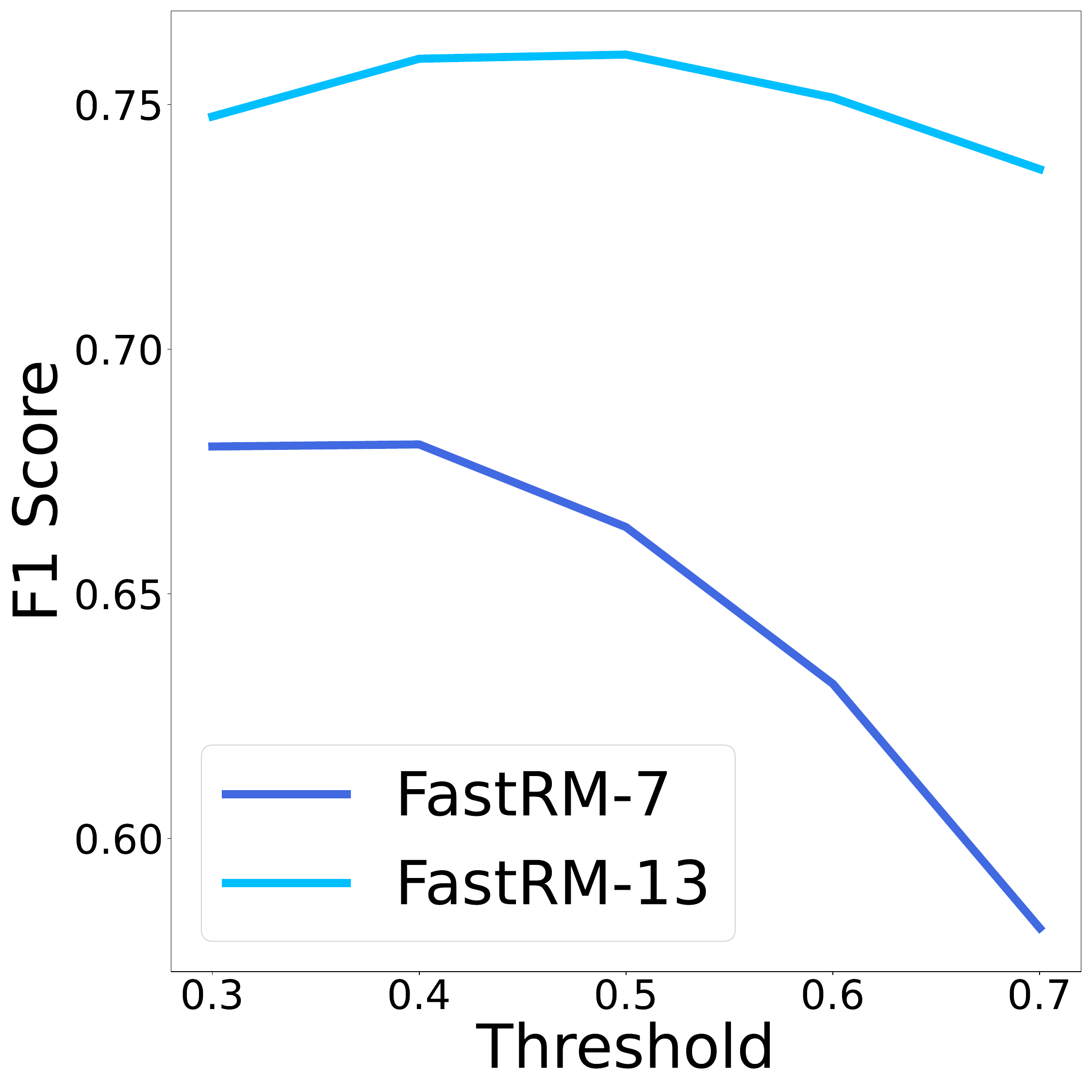}}
}
\caption{Performance of LLaVA v1.5-7B: Accuracy and F1 score of FastRM-7 and FastRM-13 across different classification thresholds.}
\end{figure}

The results in Figure~\ref{fig:accuracy} indicate that both FastRM-7 and FastRM-13 achieve the highest accuracy for the threshold of 0.5. This suggests that our model is well-calibrated and that the output probabilities are close to the true likelihood of the image token being relevant. FastRM-13 has an overall higher accuracy compared to FastRM-7. 
Since our predictions are imbalanced as described in Figure \ref{fig:Distribution-FastRM-7}, we also need to measure performance in terms of recall/precision. 
Figure \ref{fig:f1} shows how the F1 score behaves for different classification thresholds. 
For FastRM-13, the best F1-score is reached for a threshold between 0.4 and 0.5 meeting the accuracy results. For FastRM-7, as we increase the threshold beyond 0.4, our model becomes more conservative in predicting the relevant tokens, requiring higher confidence to label a token as relevant. 
Since we prioritize high recall, the optimal classification threshold should be 0.4. Indeed, we aim to ensure that our model does not miss any relevant tokens, even if it means allowing some tokens to be labeled as relevant when they are not. That way, we can make sure that the model is focusing at least on the correct areas, ensuring it attends to the right places, even if some of the attention is misplaced. This is crucial for tasks where capturing all pertinent information is essential (e.g. medical image analysis). 

To evaluate the generalizability of our proxy model, trained on the VQA training dataset, we expanded the evaluation to two additional datasets: a 10,000-sample subset from the GQA validation dataset \cite{hudson2019gqanewdatasetrealworld} and POPE \cite{li2023evaluatingobjecthallucinationlarge} which contains 9,000 samples. Table \ref{table:acc_f1_gen} presents the accuracy and F1 score per model across benchmarks for a decision threshold of 0.5. Notably, the accuracy and F1 scores are higher for GQA and POPE compared to the in-domain dataset, VQA, on which FastRM was trained. This indicates strong generalizability and demonstrates that our model does not overfit. Figure \ref{fig:vqa_gqa} in the Appendix presents the perturbation-based evaluations for FastRM-7 and shows similar behaviors to VQA, fulfilling its role in identifying regions that are crucial for the model's answers.

To demonstrate that the output generated by FastRM is indeed correlated with the model's focus on the image, we conducted positive and negative-perturbation based experiments by masking image patches based on their relevancy values. 
The underlying intuition is that masking the most relevant image patches should significantly degrade performance, and masking the least relevant ones should have minimal impact.  To enforce concise, VQA-style responses, we added to the prompt the following: \texttt{Answer the question using up to two words}. We set the decoding temperature to 0, to ensure deterministic outputs. We then incrementally removed image patches and computed VQA accuracy. In positive perturbation, image patches are removed in order of highest to lowest relevance, whereas in the negative version, they are removed from lowest to highest.

\begin{figure}[tb]
\centering     
\resizebox{\columnwidth}{!}{%
\subfigure[PP - FastRM-7]{\label{fig:pos_perturb-7b}\includegraphics[width=40mm]{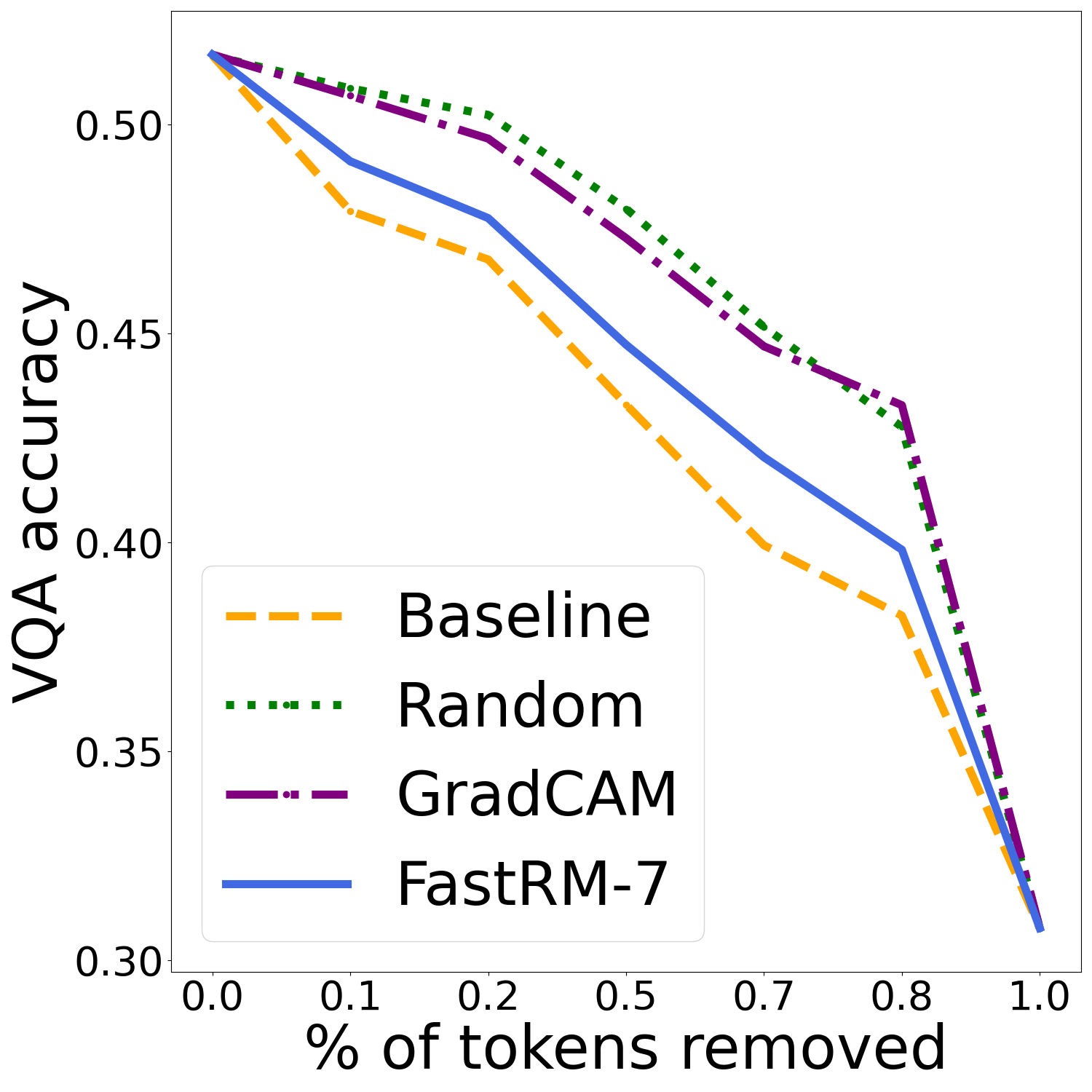}}
\subfigure[PP - FastRM-13]{\label{fig:pos_perturb-13b}\includegraphics[width=40mm]{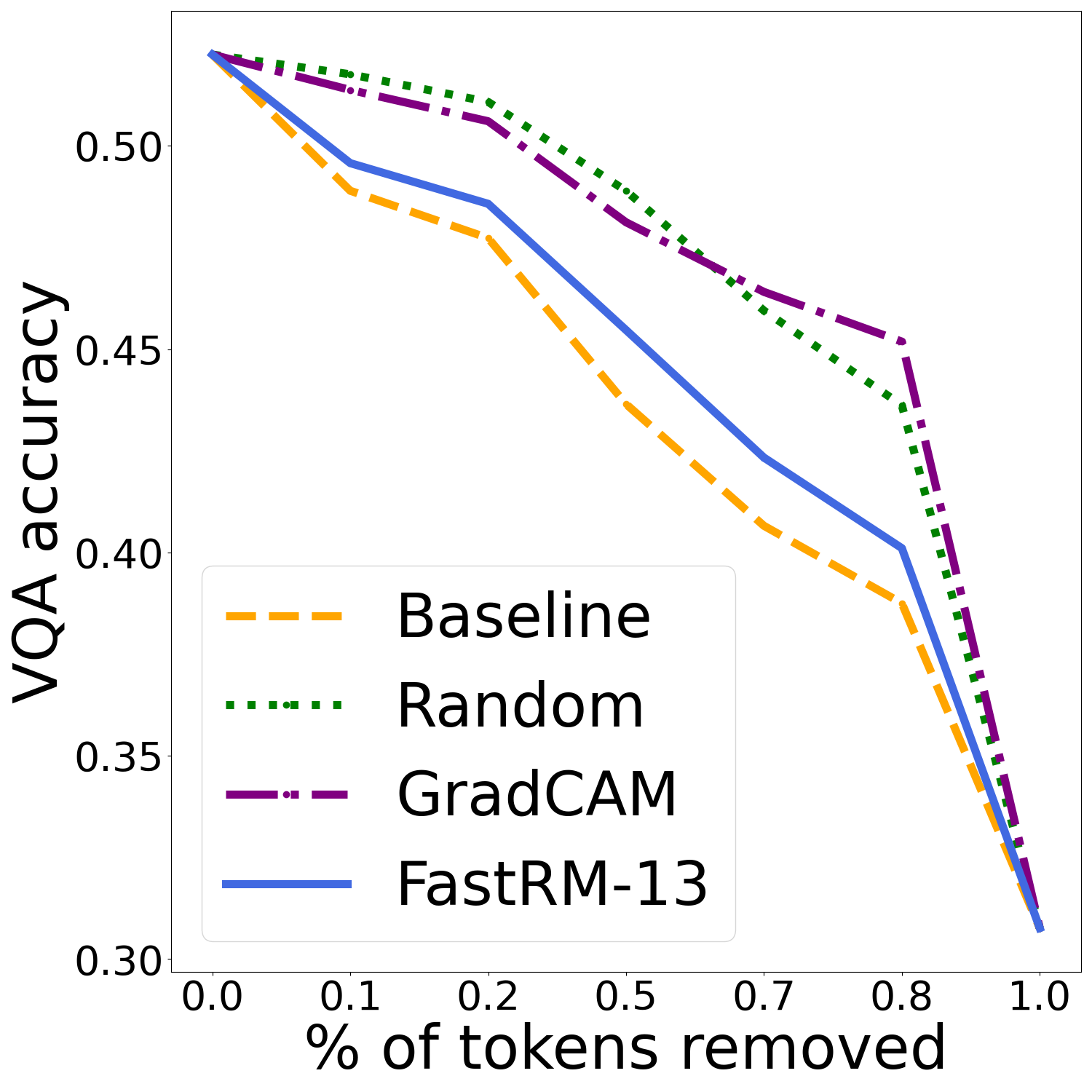}}
\subfigure[NP - FastRM-7]{\label{fig:neg_perturb-7b}\includegraphics[width=40mm]{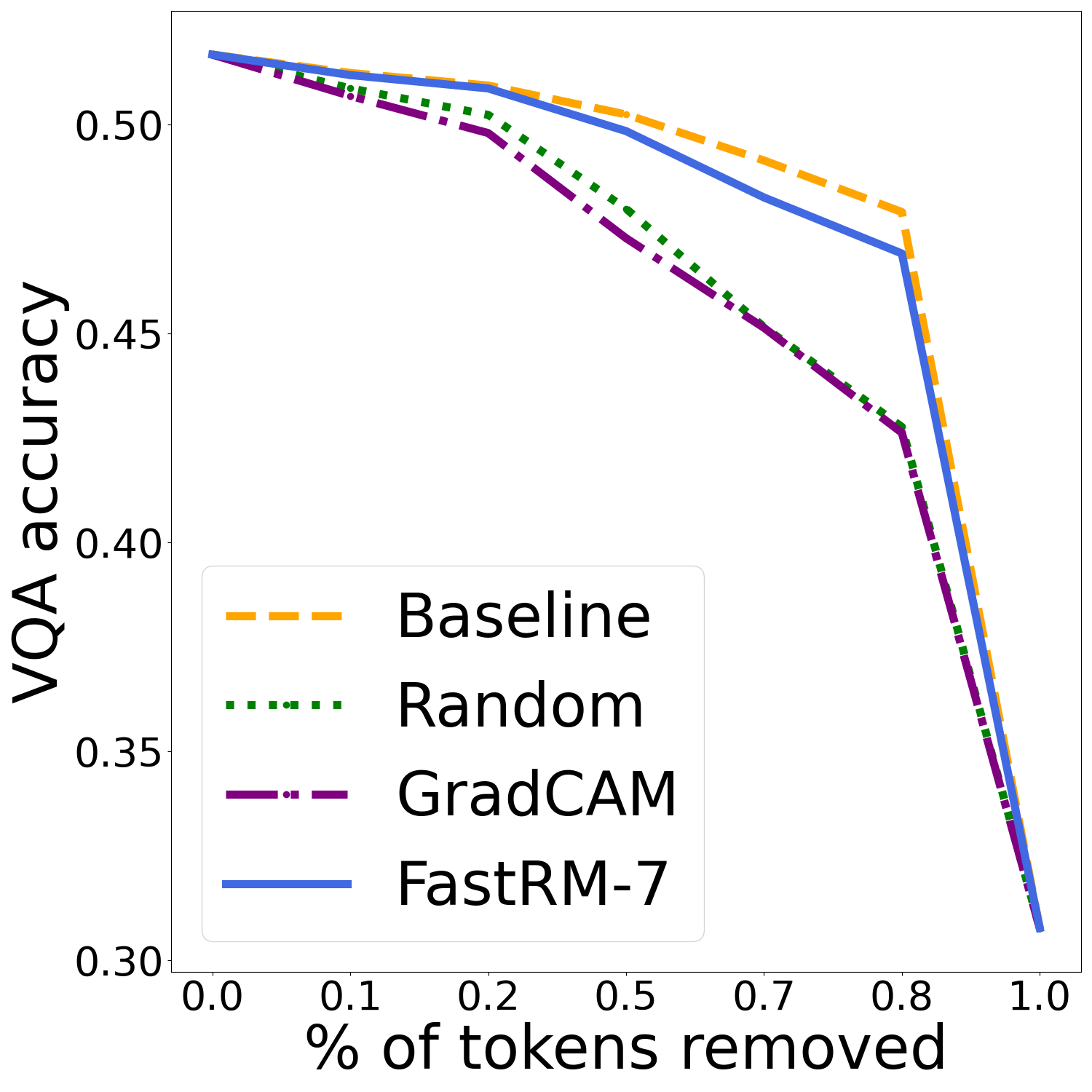}}
\subfigure[NP - FastRM-13]{\label{fig:neg_perturb-13b}\includegraphics[width=40mm]{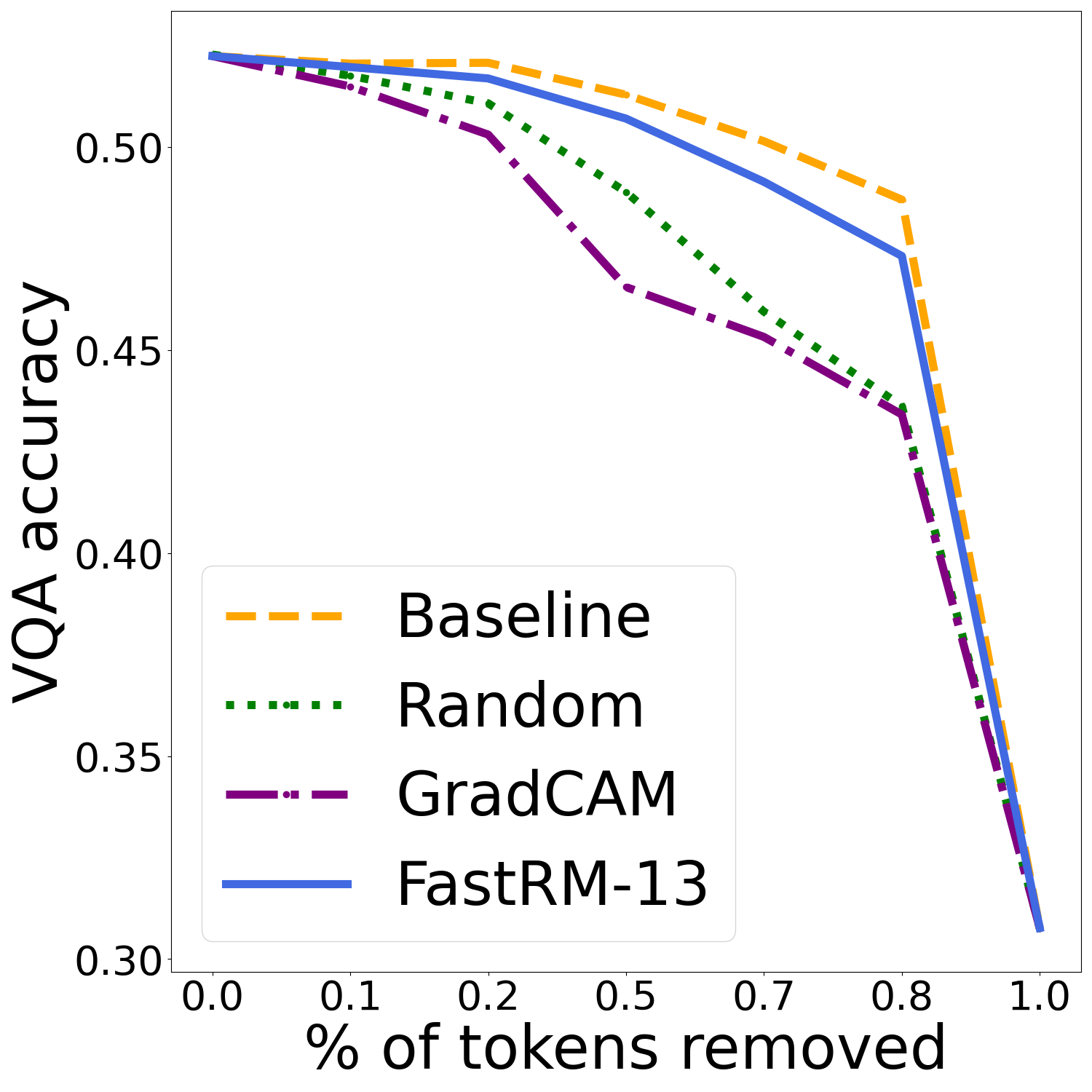}}
}
\caption{Perturbation-based evaluation. For positive perturbation, PP (smaller AUC is better). For negative perturbation, NP (larger AUC is better). }
\label{fig:perturb}
\end{figure}

\begin{figure}[tb]
\centering     
\subfigure[Labeling threshold]{\label{fig:pos_ablations_threshold}\includegraphics[width=40mm]{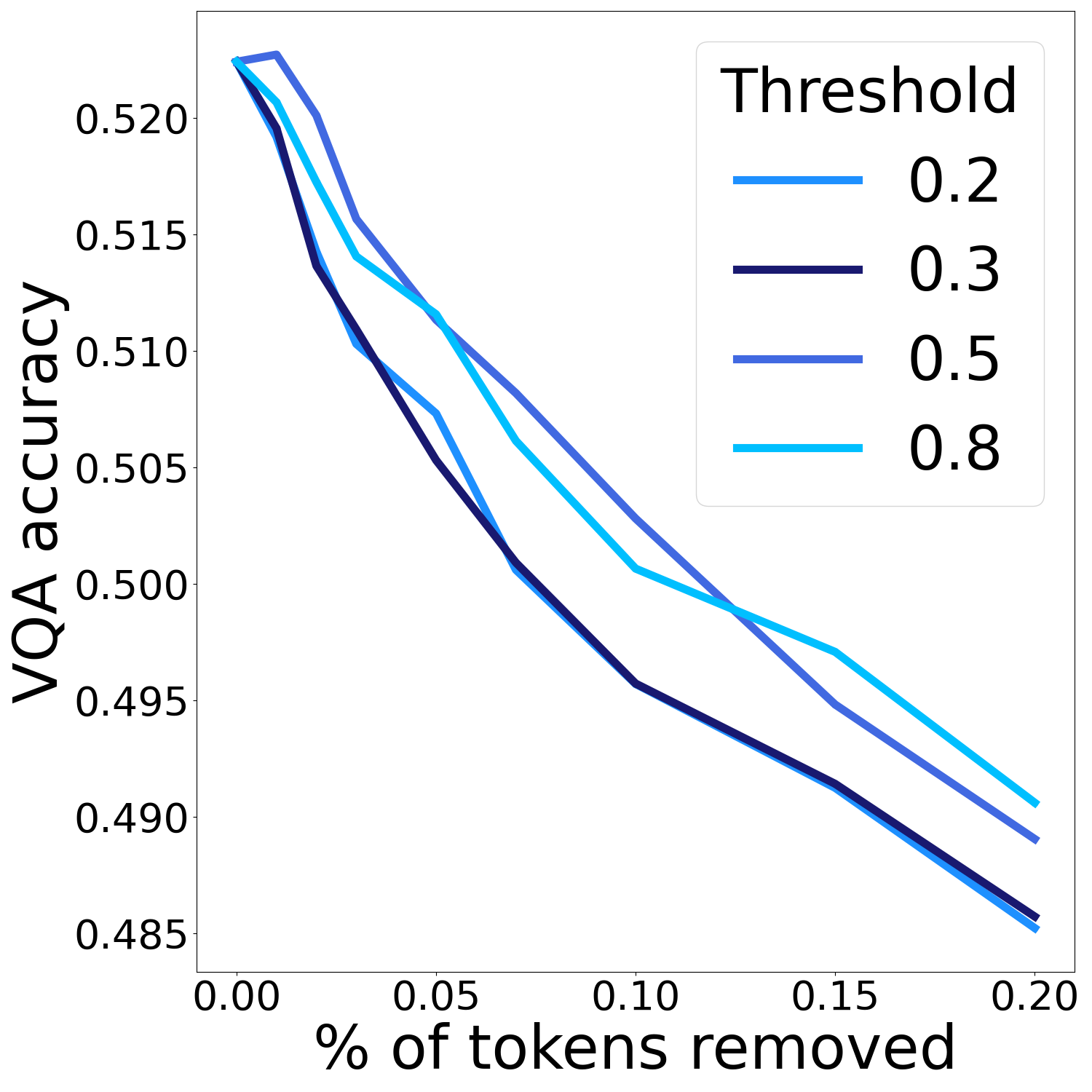}}
\subfigure[Training dataset size]{\label{fig:data_size_ablation_pertubation-7b}\includegraphics[width=40mm]{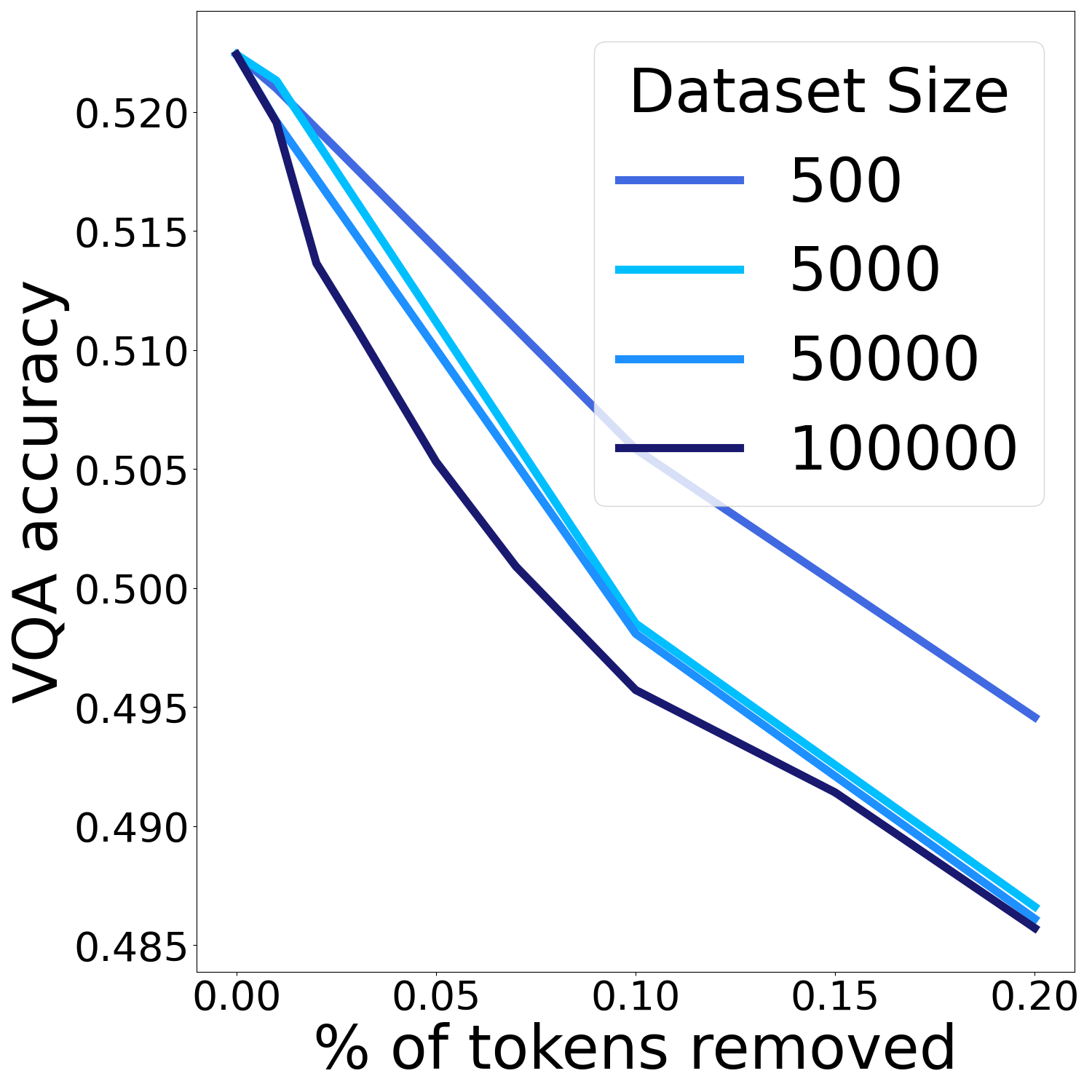}}
\subfigure[Number of training steps]{\label{fig:steps_ablation_pertubation-7bfig:b}\includegraphics[width=40mm]{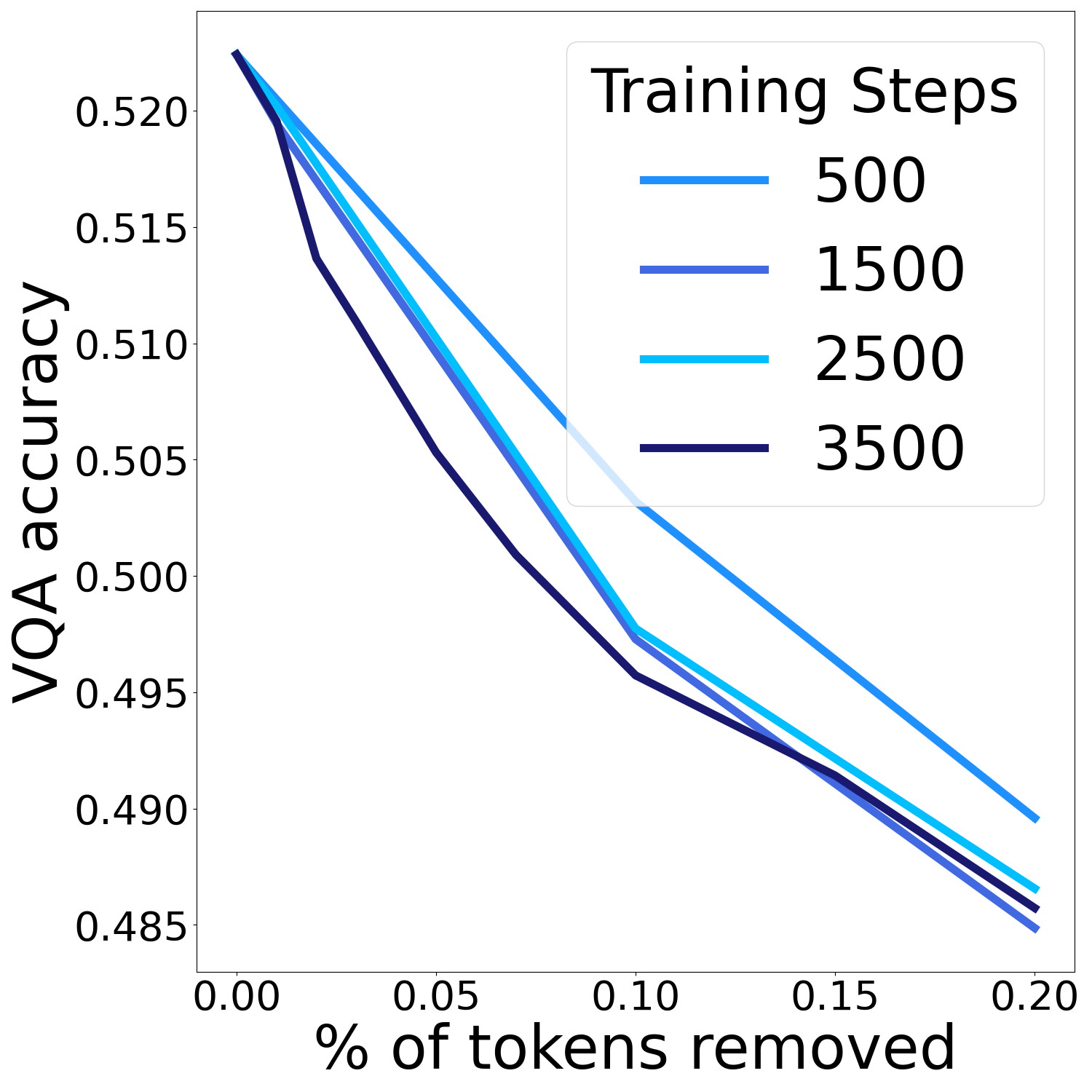}}
\caption{Ablations studies for FastRM-13. The figure shows how the labeling threshold, the training data size and the number of training steps affect the PP-based evaluation. Smaller AUC is better }
\label{fig:perturb2}
\end{figure}

We compared our approach with the following methods: random removal of image patches, GradCAM as introduced by ~\cite{selvaraju2017grad}, and relevancy maps computed based on attentions and their gradients as presented by ~\cite{chefer2021transformerinterpretabilityattentionvisualization}, referred to as the "Baseline" throughout the paper. For random selection, we performed five experiments and reported the average performance.
Figures ~\ref{fig:pos_perturb-7b} and \ref{fig:pos_perturb-13b}  illustrate the results of masking different percentages of image patches. As we progressively mask image tokens selected by different methods, our method leads to a large degradation in performance for the positive perturbations. This indicates that FastRM selects the most relevant image patches more effectively compared to the other methods. 
Our method consistently results in a greater performance drop than the random selection and GradCAM methods. 
However, we notice a sensitively slower performance drop than the baseline. This slightly lower performance is expected when dealing with model distillation and introduces a trade-off between memory/speed and accuracy. 
Similarly, we also performed negative perturbation-based evaluations, in which masking the least relevant image patches is expected to have a minimal effect on accuracy. Figures \ref{fig:neg_perturb-7b} and \ref{fig:neg_perturb-13b} show a comparison between FastRM, GradCAM, baseline, and random removal. We observe that as we mask the least probable relevant patches of the image, FastRM's accuracy decrease is comparable with that of the baseline and slower than both random and GradCAM.  Indicating FastRM is robust to the removal of less probable relevant patches, and on par with the baseline.

\begin{table}[tb!]
\centering
\scriptsize
\resizebox{0.8\columnwidth}{!}{%
\begin{tabular}{|c|c|c|c||c|c|c|c|}
\hline
  & \multicolumn{3}{|c||}{Accuracy} & \multicolumn{3}{|c|}{F1 Score} \\ \hline
  & VQA & GQA & POPE & VQA & GQA & POPE \\ \hline
FastRM7 & 0.981 & 0.999 & 0.99 & 0.57 & 0.66 & 0.63 \\ \hline
FastRM13 & 0.993 & 0.996 & 0.998 & 0.72 & 0.81 & 0.84 \\ \hline
\end{tabular}%
}
\caption{Accuracies and F1 Scores for FastRM7 and FastRM13 across benchmarks.}
\label{table:acc_f1_gen}
\end{table}



\begin{figure}[tb]

    \begin{minipage}{\textwidth}
        \centering
        \includegraphics[width=\textwidth]{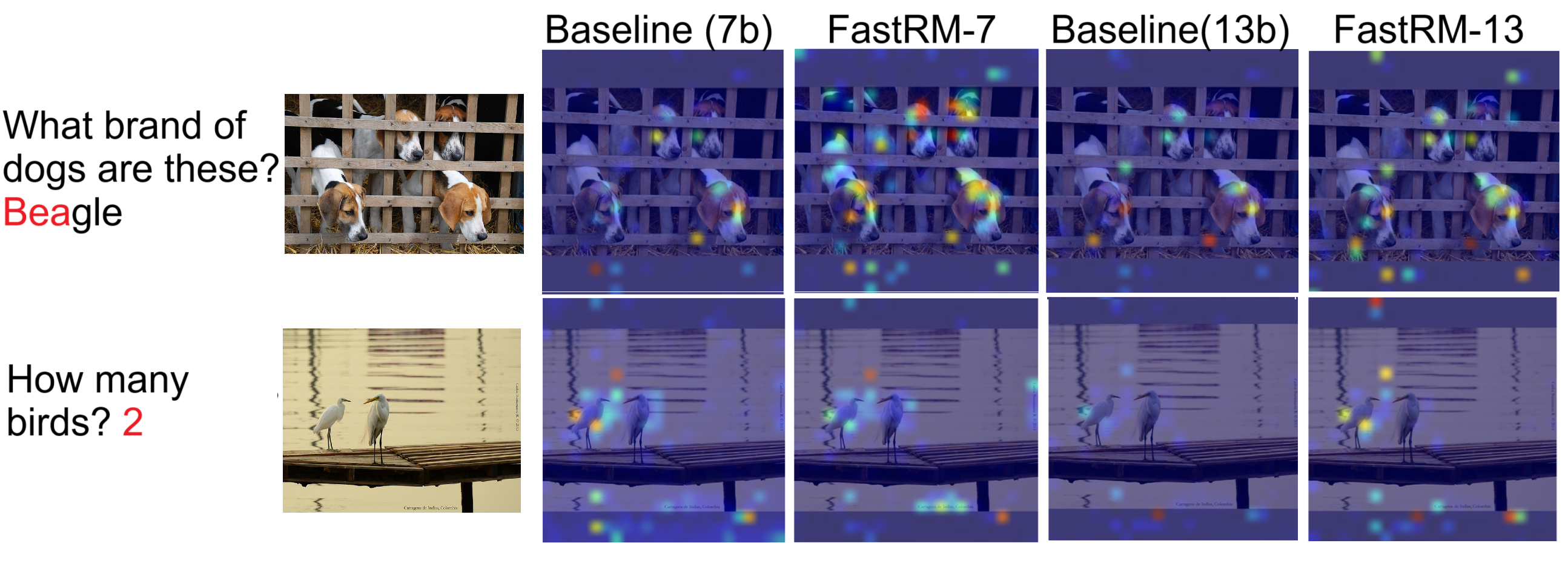}
        \caption{Qualitative comparison between the baseline, FastRM-7 and FastRM-13. The relevancy maps corresponding to the token highlighted in red from the answer.}
        \label{fig:quali}
    \end{minipage}
\end{figure}

\paragraph{Qualitative}
We see a clear correlation between the sigmoid activated FastRM outputs and the baseline,  as shown in Figure \ref{fig:quali} for both LLaVAv1.5-7B and LLaVAv1.5-13B.
This suggests that the model has learned to predict the intensity of the relevancy scores on its own, likely capturing the underlying patterns of relevance in a way that goes beyond the binary values. 
This supports our initial hypothesis that the last hidden states contain enough information to capture the patterns of the relevancy map. 

\begin{figure}[tb]
\centering     
\resizebox{\columnwidth}{!}{%
\subfigure[Latency (in sec) according to the number of output tokens]{\label{fig:latency_gain}\includegraphics[width=60mm]{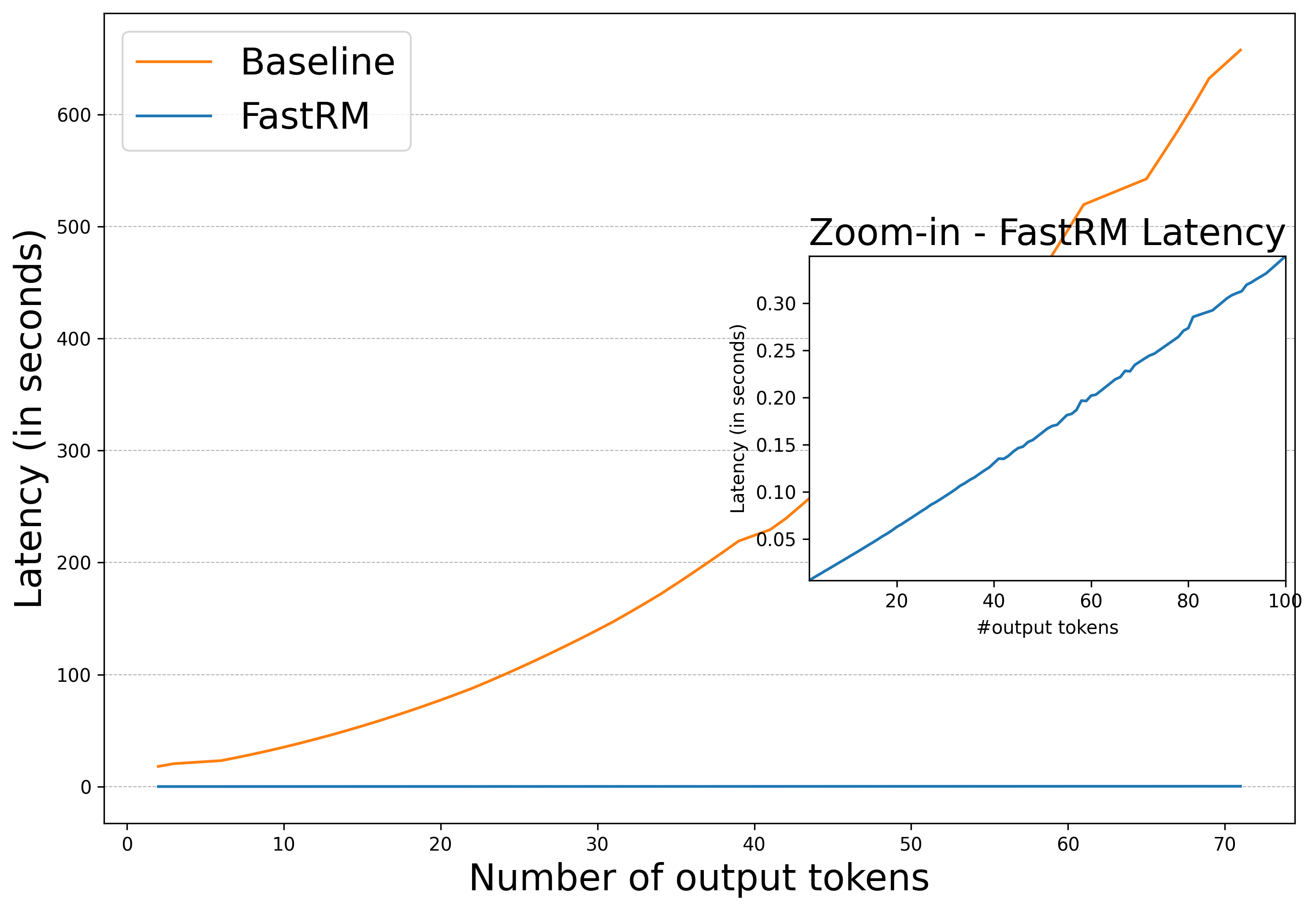}}
\subfigure[Memory (in Gb) according to the number of output tokens]{\label{fig:all_mem_gain}\includegraphics[width=60mm]{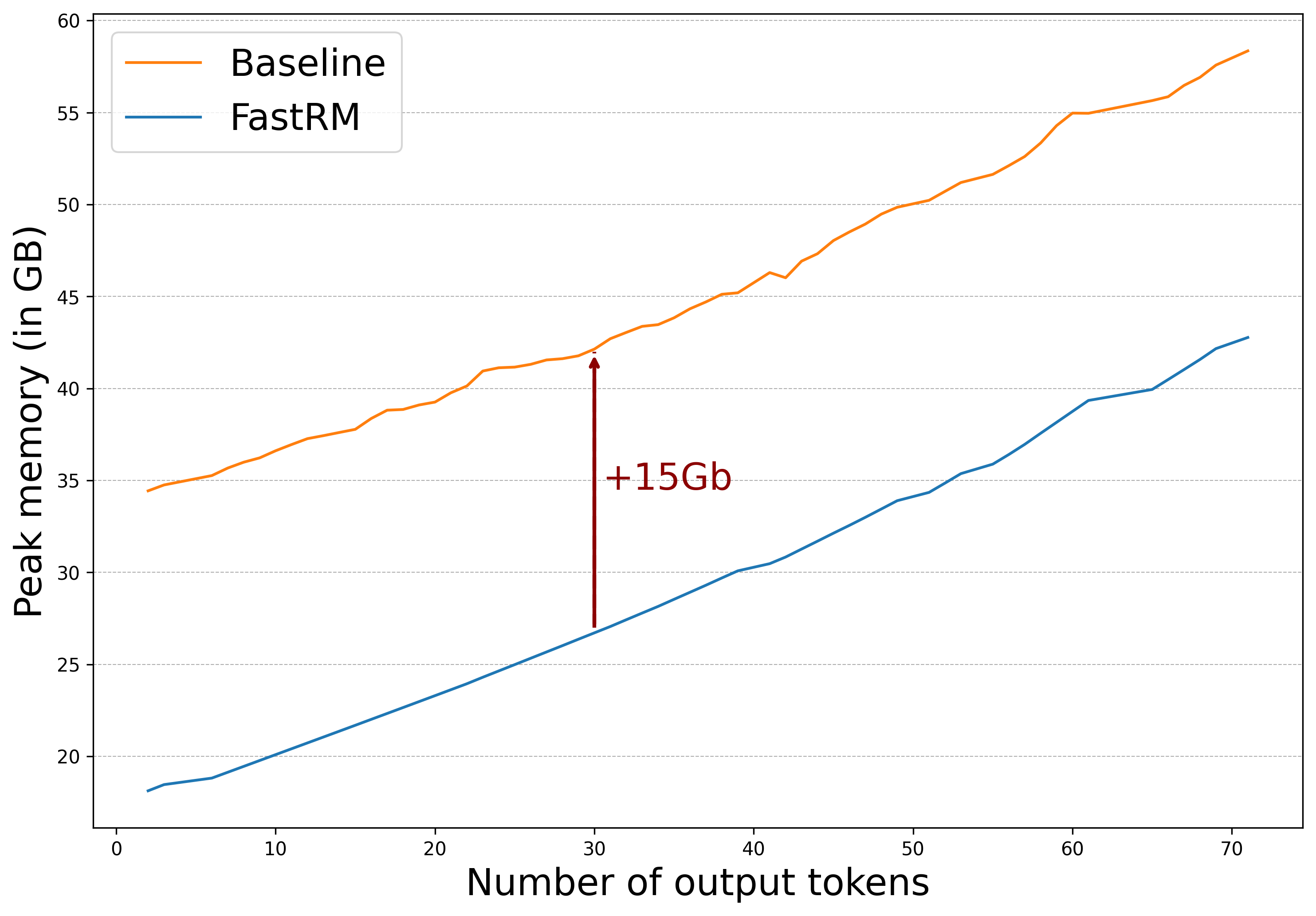}}
}
\caption{Latency and memory footprint for LLaVA v1.5-7B}
\end{figure}

\paragraph{Latency and Memory Efficiency}
We evaluated the gain in latency and memory footprint on 1000 samples from VQA with a maximum of generated tokens of 100. We first measured memory footprint and latency using the original computation of the relevancy maps as stated in \cite{chefergame} using LLavA-v1.5-7b and we compared it with FastRM-7. 
Figure \ref{fig:latency_gain} shows an exponential increase in latency as the number of output tokens increases while our method shows a minimal increase. 
Specifically, generating 10 and 100 output tokens takes respectively around 15 seconds and 14 minutes with the baseline method, while FastRM takes just 0.35 seconds for 100 output tokens. Considering that most queries for VQA require between 10 to 30 tokens to answer, as shown in Figure \ref{fig:histo_len}, FastRM significantly accelerates the process, making it a more efficient solution for typical output lengths. VQA involves short questions needing brief answers, but for benchmarks requiring longer answers, the computation time difference will be even greater.
On average, the latency for computing the relevancy map using the original calculations is 620 times greater than with our method, representing a 99.8\% reduction in time.
Figure \ref{fig:all_mem_gain} also demonstrates a 44\% memory reduction on average compared to the baseline during the entire generation phase.
All the experiments described in this section were conducted on one NVIDIA A100.

\paragraph{Ablation studies}
We conducted ablation studies to assess the impact of labeling threshold, dataset size, and training steps for both FastRM-7 and FastRM-13 performance. We performed positive perturbation experiments based on their relevancy values probabilities and evaluated the VQA accuracy. We focused on the first 20\% of token removal, as this is where we expect to observe the steeper drop. We focus on ablations for FastRM-13, as FastRM-7 shows less significant differences across the ablated metrics.
 Figure \ref{fig:pos_ablations_threshold} demonstrates the effect of different labeling thresholds on performance, revealing that a threshold of 0.3 yields the best performance.
We trained FastRM-13 with different dataset sizes of 500, 5k, 50k, and 100k samples. \ref{fig:data_size_ablation_pertubation-7b} indicates that the larger training set lead to better performance.
Finally, \ref{fig:steps_ablation_pertubation-7bfig:b} shows that longer training leads to better performance. In the latest ablations, FastRM-7 exhibits a similar behavior.

\paragraph{Quantifying model confidence}



We hypothesize that a more concentrated model is more likely to generate a correct answer, as it is probably focusing on the relevant area.
We quantified this intuition by computing the entropy of the relevancy maps produced by FastRM.
On a subset of 10k samples from the VQA validation dataset, where LLaVA-7b had an accuracy of 74\%, we generated relevancy maps using FastRM-7 and computed their entropy. To estimate the probability density function of the entropy values, we applied Kernel Density Estimation (KDE). Figure \ref{fig:histo_uncertainty} displays the histogram of the resulting values, highlighting the distribution differences when accuracy is 0 and 1. Once the KDE has been applied, we can compute the entropy as described in 
$H(X) = - \sum_{i=1}^{N} p(x_i) \log p(x_i)
$ where $x_i$ are the relevancy scores and $N$ the number of patches.

The threshold (red dashed line in Figure \ref{fig:histo_uncertainty}) was determined as the arithmetic mean of the means of both distributions. 
According to the Figure, when the entropy of the relevancy map is below the threshold (representing 63\% of the set), the model is highly likely to be correct. However, when the entropy exceeds this threshold (37\% of the set), predicting the expected result becomes more challenging. 
To validate this hypothesis, we measured the false positives (FP) and true positives (TP) in this region across 10k samples (different samples from those used to generate the threshold). In this situation, a TP means that we expected the answer to be correct and it is, while FP means that we expected it to be correct when it is not. In the region below the threshold, we hypothesize that the model accuracy will be 1. We observe 84\% of true positives (we expected the model accuracy to be 1 and it is) against 16\% of false positives (we expected the model accuracy to be 1 and it is not). This result indicates that below the threshold, we can have greater confidence in the model's correctness. However, in the region above the threshold, we find 59\% of false negatives and 41\% of true negatives. This suggests that when the entropy is in this range, the model has nearly an equal chance of making a correct or incorrect prediction. We conducted the same experiments on the GQA and POPE benchmarks to assess the method's generalizability. We found that below the thresholds, set using the same technique, we can be confident of the model's correctness in 83\% of cases for GQA and 86\% for POPE.

\section{Conclusion, Limitations and Future Work}
In this work we introduced FastRM, a lightweight model distilled from an explainable metric that improves memory efficiency and reduces latency, enabling real-time model decision qualitative interpretation. FastRM also provides a way to measure the confidence of a model correctness, crucial for high-stakes scenarios like healthcare and autonomous driving.
 The limitations include the model being designed solely to provide a probability of a token's relevance, rather than offering exact relevancy maps.
Future work could explore applying our approach to other architectures beyond LLaVA models and to other explainability methods that produce saliency maps. Moreover, integrating FastRM during LVLM training would lead to improved vision-language grounding, while its low latency and memory footprint enable seamless feedback integration.

\clearpage

\bibliography{iclr2025_conference}
\bibliographystyle{iclr2025_conference}
\clearpage

\appendix
\section{Appendix}
\subsection{Additional Quantitative Results}
\label{sec:Additional_Quantitative}

\begin{figure}[H]
\centering     
\resizebox{\columnwidth}{!}{%
\subfigure[PP on POPE]{\label{fig:pp_pope}\includegraphics[width=30mm]{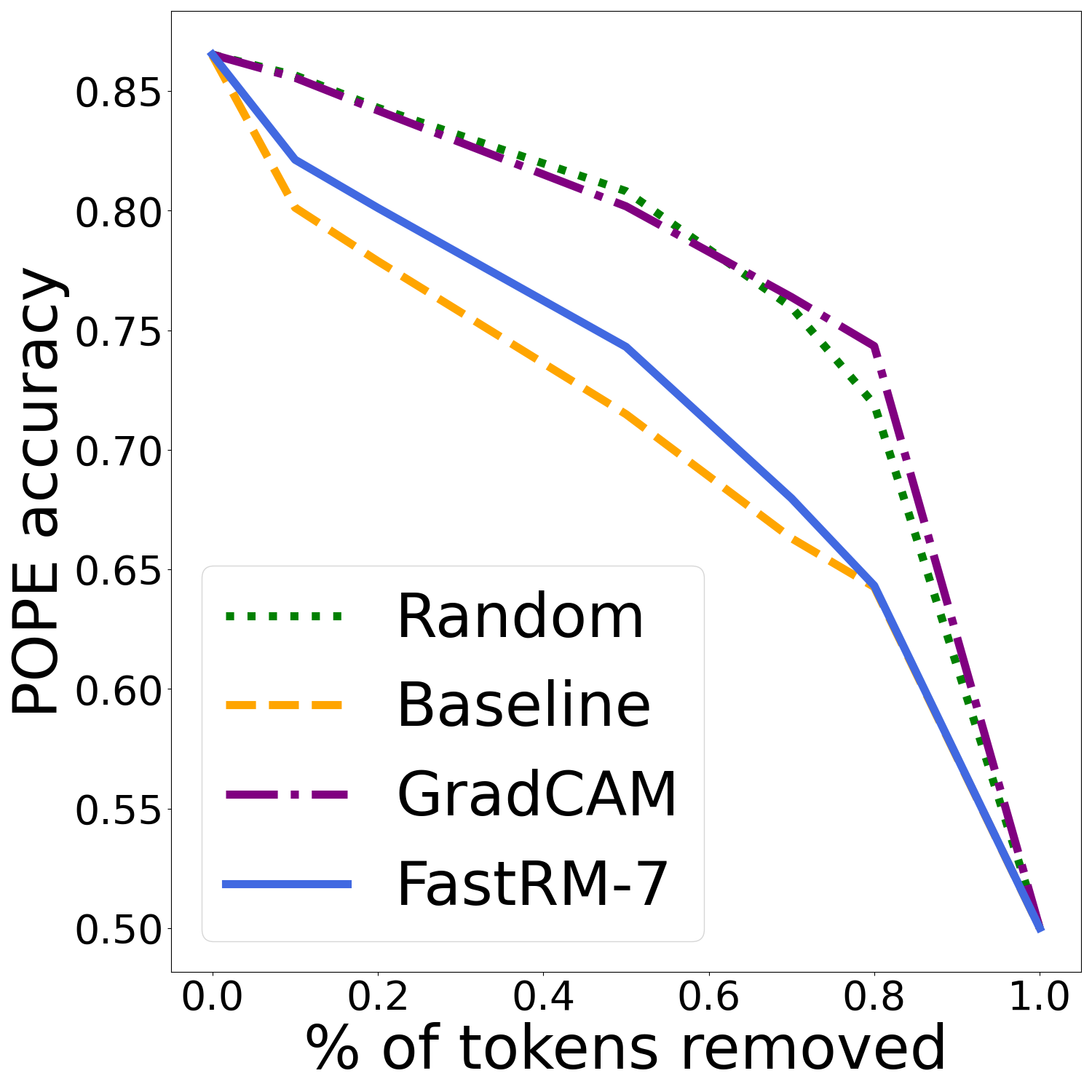}}
\subfigure[PP on GQA]{\label{fig:pp_gqa}\includegraphics[width=30mm]{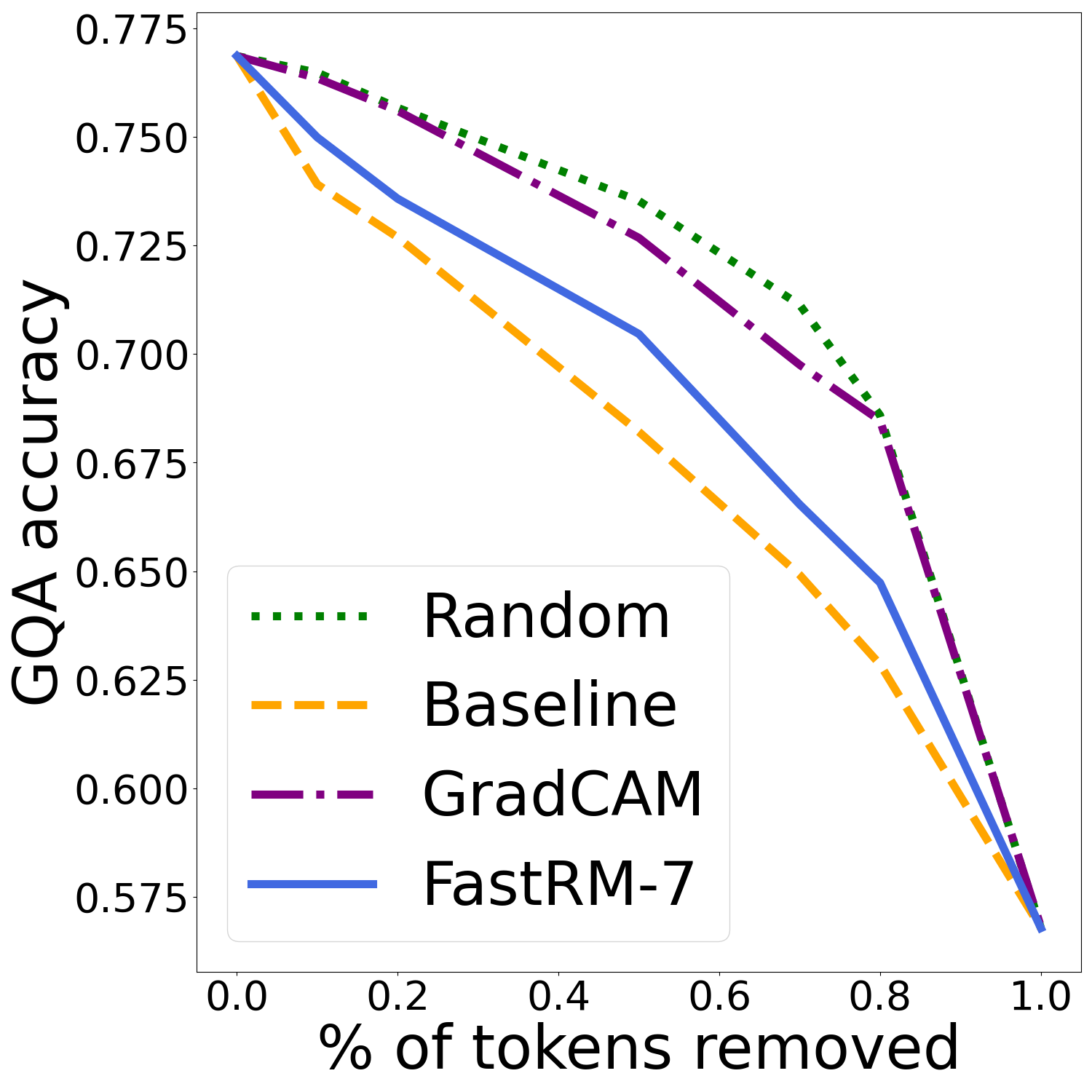}}
\subfigure[NP on POPE]{\label{fig:np_pope}\includegraphics[width=30mm]{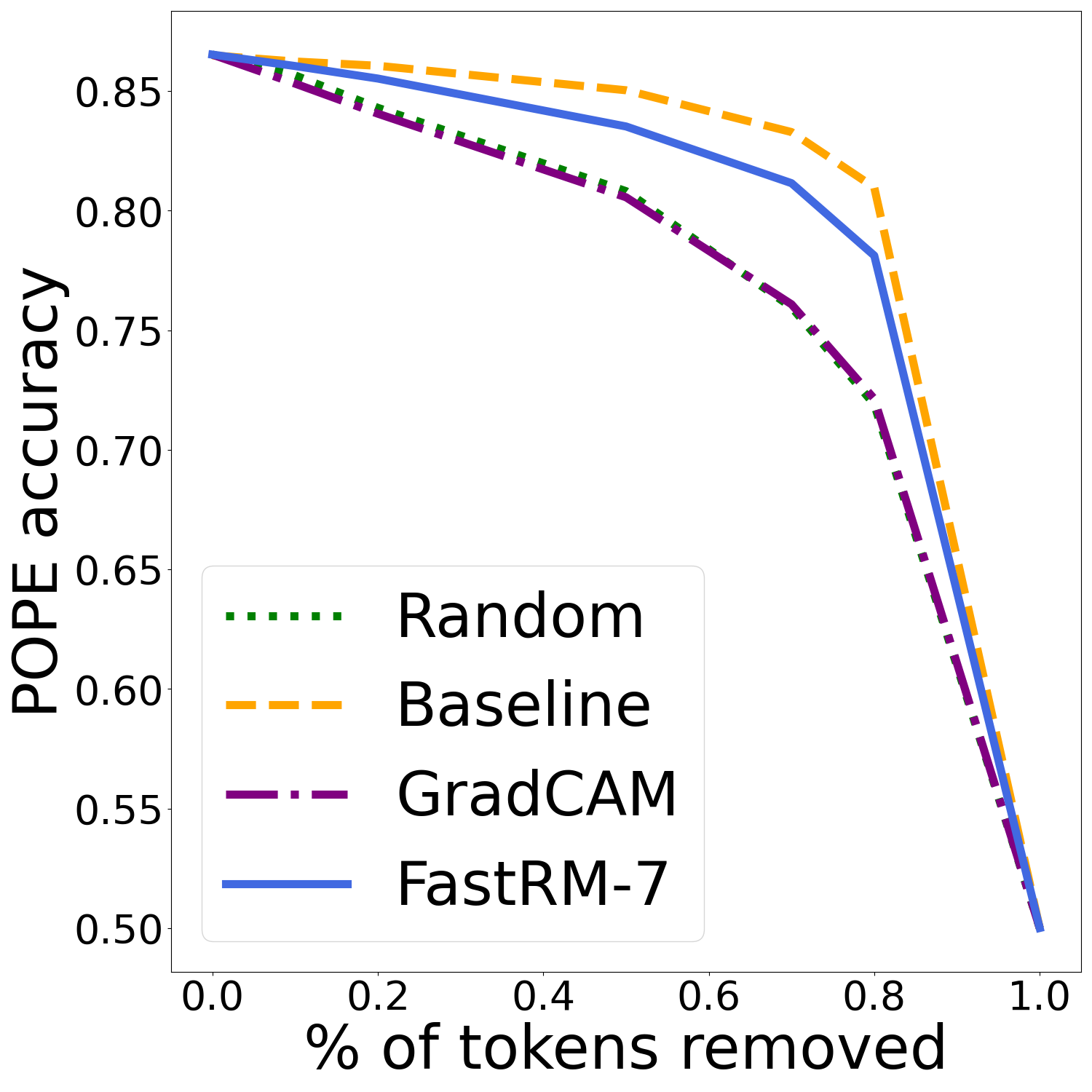}}
\subfigure[NP on GQA]{\label{fig:np_gqa}\includegraphics[width=30mm]{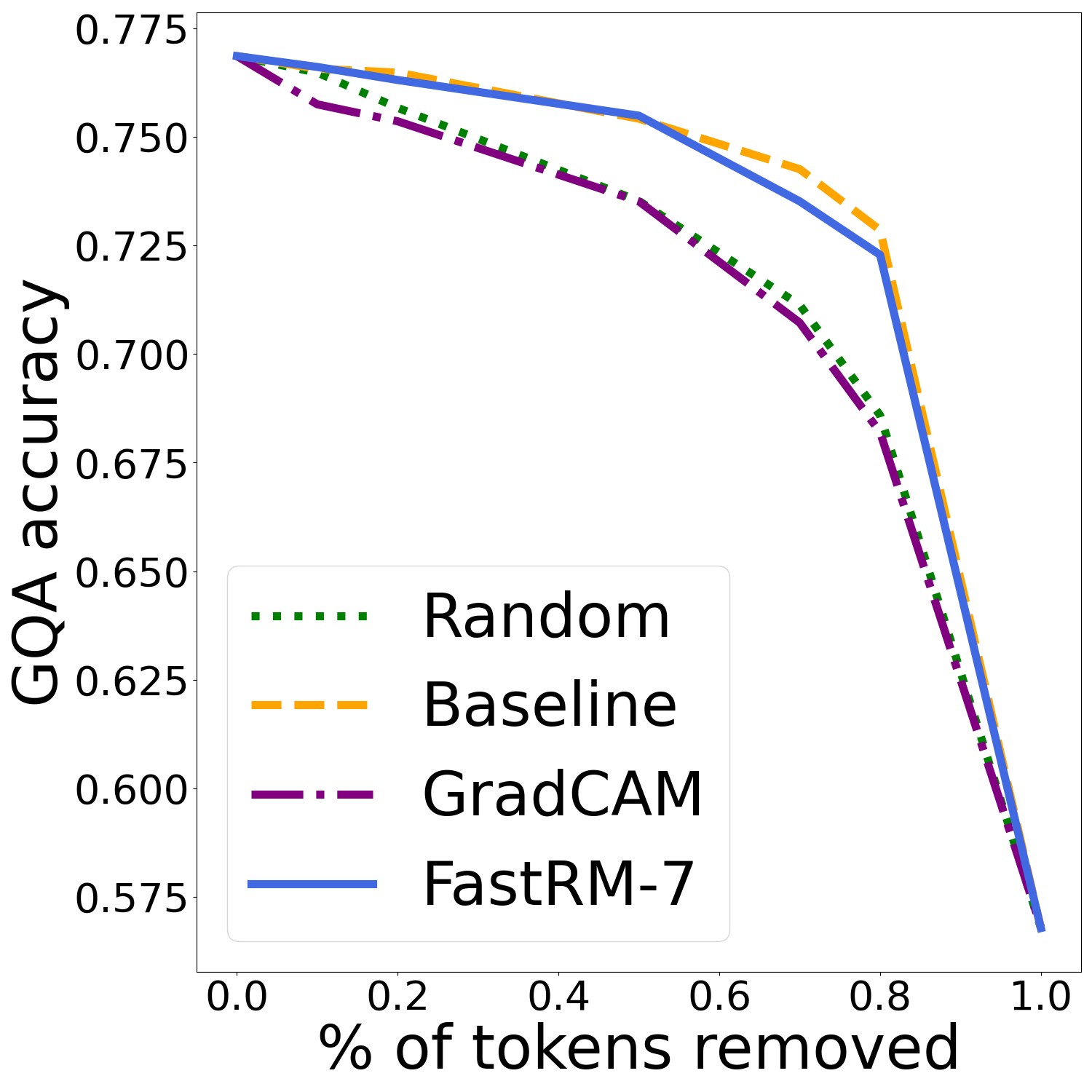}}
}
\caption{Perturbation-based evaluation of FastRM-7 on POPE and GQA.  }

\label{fig:vqa_gqa}
\end{figure}

To assess the generalizability of our proxy model, which was trained on the VQA training dataset, we tested the FastRM proxy on two additional datasets: a 10,000-sample subset of the GQA validation dataset and POPE, which consists of 9,000 samples.
We compare the performance of FastRM to the baseline model, random removal of image patches, and GradCAM, as shown in Figure~\ref{fig:vqa_gqa}.

\end{document}

%% file: math_commands.tex
\usepackage{amsmath,amsfonts,bm}

\def\eqref#1{equation~\ref{#1}}

\def\1{\bm{1}}

\DeclareMathAlphabet{\mathsfit}{\encodingdefault}{\sfdefault}{m}{sl}
\SetMathAlphabet{\mathsfit}{bold}{\encodingdefault}{\sfdefault}{bx}{n}

